\documentclass[letterpaper]{article}
\usepackage{uai2019}
\usepackage[margin=1in]{geometry}

\usepackage{times}

\usepackage[T1]{fontenc}

\usepackage{tgtermes}
\usepackage{amsmath}
\usepackage{scalefnt,letltxmacro}
\LetLtxMacro{\oldtextsc}{\textsc}
\renewcommand{\textsc}[1]{\oldtextsc{\scalefont{1.10}#1}}
\usepackage[scaled=0.92]{PTSans}

\usepackage[usenames,dvipsnames]{xcolor}
\definecolor{shadecolor}{gray}{0.9}

\usepackage[final,expansion=alltext]{microtype}
\usepackage[english]{babel}
\usepackage[parfill]{parskip}
\usepackage{afterpage}
\usepackage{framed}
\usepackage{nicefrac}

\usepackage{lineno}

\usepackage{ragged2e}

\usepackage{graphicx}
\usepackage[labelfont=bf]{caption}
\usepackage{subfig}

\usepackage{booktabs}
\usepackage{tabu}

\usepackage{listings}
\usepackage{fancyvrb}
\fvset{fontsize=\normalsize}

\usepackage[colorlinks,linktoc=all]{hyperref}
\usepackage[all]{hypcap}
\hypersetup{citecolor=Violet}
\hypersetup{linkcolor=black}
\hypersetup{urlcolor=MidnightBlue}

\usepackage[nameinlink]{cleveref}

\usepackage
[acronym,smallcaps,nowarn,section,nogroupskip,nonumberlist]{glossaries}
\glsdisablehyper{}
\crefname{appendix}{supplement}{}

\usepackage{listings}
\lstdefinestyle{alp_style}{
	commentstyle=\color{OliveGreen},
	numberstyle=\tiny\color{black!60},
	stringstyle=\color{BrickRed},
	basicstyle=\ttfamily\scriptsize,
	breakatwhitespace=false,
	breaklines=true,
	captionpos=b,
	keepspaces=true,
	numbers=none,
	numbersep=5pt,
	showspaces=false,
	showstringspaces=false,
	showtabs=false,
	tabsize=2
}
\usepackage{amsmath, amsfonts, amssymb}
\usepackage[usenames,dvipsnames]{xcolor}
\usepackage{colortbl}
\usepackage{dsfont}
\usepackage{float}
\usepackage{tikz}
\usepackage{multirow}

\usepackage[round]{natbib}

\usepackage{algorithm}
\usepackage{algpseudocode}
\usepackage{courier} 
\usepackage{adjustbox}

\newcommand{\bX}{\boldsymbol{X}}
\newcommand{\bx}{\boldsymbol{x}}
\newcommand{\boldf}{\boldsymbol{f}}
\newcommand{\by}{\boldsymbol{y}}
\newcommand{\bn}{\boldsymbol{n}}
\newcommand{\bu}{\boldsymbol{u}}
\newcommand{\bgamma}{\boldsymbol{\gamma}}
\newcommand{\bomega}{\boldsymbol{\omega}}
\newcommand{\blambda}{\boldsymbol{\lambda}}

\newcommand{\bmu}{\boldsymbol{\mu}}
\newcommand{\Po}{\mathrm{Po}}
\newcommand{\PG}{\mathrm{PG}}
\newcommand{\GP}{\mathrm{GP}}
\newcommand{\Ga}{\mathrm{Ga}}

\newcommand{\diag}{\mathrm{diag}}

\newcommand{\methodname}{Conjugate multi-class Gaussian process classification}
\newcommand{\namelikelihood}{logistic-softmax}

\newcommand{\expec}[2]{\mathbb{E}_{#1}\left[#2\right]}

\newcommand{\uptoconst}{\mathrel{\overset{\makebox[0pt]{\mbox{\normalfont\small\sffamily c}}}{=}}}

\newcommand{\LSM}{\textsc{lsm}}
\newcommand{\RM}{\textsc{rm}}
\newcommand{\HS}{\textsc{hs}}
\newcommand\scale[2]{%
  \ifnum#1>#2
    $#1 > #2$
  \else
    \ifnum#1<0
      $#1 < 0$
    \else
      \ifnum#2<0
        $#2 < 0$
      \else
        \tikz{%
        \ifx#20
        \else
          \foreach \i in {1,...,#2} {
            \filldraw[black!20] (\i ex,0) circle (0.4ex);
          };
        \fi
        \ifx#10
        \else
          \foreach \i in {1,...,#1} {
            \filldraw[black] (\i ex,0) circle (0.4ex);
          };
        \fi
        }
      \fi
    \fi
  \fi
}

\title{Multi-Class Gaussian Process Classification Made Conjugate:\\ Efficient Inference via Data Augmentation}

\author{
{\bf Th\'eo Galy-Fajou
\thanks{\hspace{0.14cm}Equal contribution. Contact: galy-fajou@tu-berlin.de.}
} \\
TU Berlin\\
Germany
\And
{\bf Florian Wenzel\hspace{0.102cm}\footnotemark[1]} \\
TU Kaiserslautern\\
Germany
\And
{\bf Christian Donner}  \\
TU Berlin\\
Germany
\And
{\bf Manfred Opper}   \\
TU Berlin\\
Germany
}

\begin{document}

\maketitle
%

%



\begin{abstract}
We propose a new scalable multi-class Gaussian process classification approach building on a novel modified softmax likelihood function.
The new likelihood has two benefits: it leads to well-calibrated uncertainty estimates and allows for an efficient latent variable augmentation.
The augmented model has the advantage that it is conditionally conjugate leading to a fast variational inference method via block coordinate ascent updates.
Previous approaches suffered from a trade-off between uncertainty calibration and speed. Our experiments show that our method leads to well-calibrated uncertainty estimates and competitive predictive performance while being up to two orders faster than the state of the art.
\end{abstract}

\section{Introduction}



In real-world decision making systems, it is important that classification methods do not only provide accurate predictions, but also indicate when they are likely to be incorrect.
Calibrated confidence estimates are important in many application domains such as self driving cars \citep{journals/corr/BojarskiTDFFGJM16}, medical diagnosis \citep{DBLP:conf/kdd/CaruanaLGKSE15} and speech recognition \citep{DBLP:journals/corr/XiongDHSSSYZ16a}.

In multi-class classification tasks, modern deep neural networks achieve state-of-the-art accuracies but often suffer from bad calibration \citep{DBLP:conf/icml/GuoPSW17}.
Gaussian process (GP) models provide an attractive alternative approach to multi-class classification problems.

Due to the Bayesian treatment of uncertainty, GPs have the advantage of leading to well-calibrated uncertainty estimates \citep{Williams98bayesianclassification, Rasmussen:2005:GPM:1162254}.
Furthermore, GP models become more expressive as the number of data points grows and allow for incorporating prior knowledge by using different kernel functions. However, inference in multi-class GP classification models is challenging.

In the easier setting of 
{\it binary} classification, GPs
can be applied to big datasets using variational inference methods \citep{Hensman2015, Wenzel_PG}.
This is possible because the expectation of generic log-likelihoods in the variational objective
(the so-called ELBO) over the variational distribution (typically a Gaussian)
reduces to univariate integrals which can be performed in an efficient way by using numerical quadrature methods. The optimization of the variational objective can then be achieved by stochastic gradient methods involving mini-batches. A further speedup of such methods is possible by the application of natural gradient techniques \citep{DBLP:conf/aistats/SalimbeniEH18}.

\begin{figure}[!ht]
	\centering
	\includegraphics[width=0.9\columnwidth]{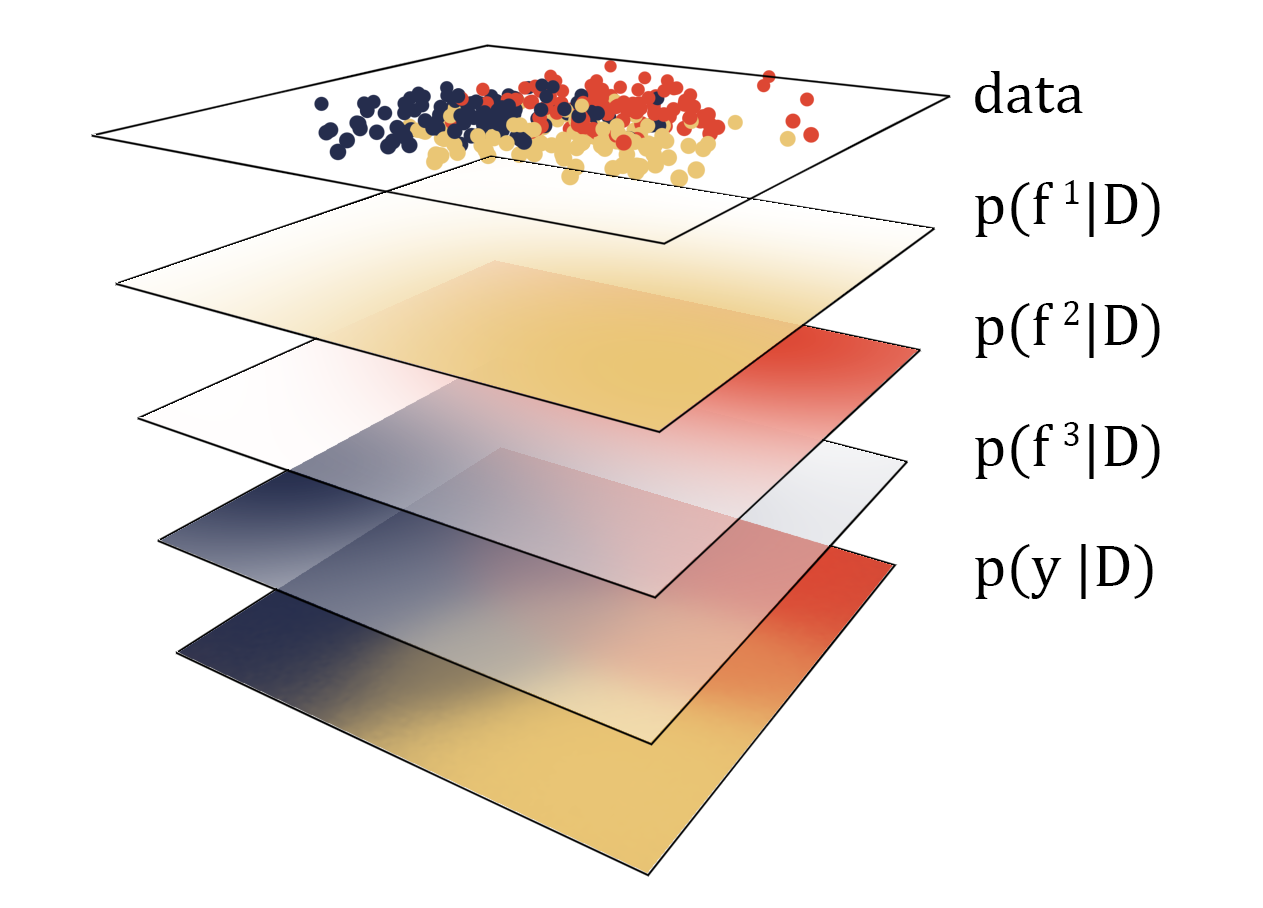}
	\caption{In a GP multi-class classification model, each class density is modeled by an individual GP $p(f^c|D)$. For predictions $p(y|D)$, the latent GPs are marginalized out.}
	\label{fig:3d}
\end{figure}

The {\it multi-class} problem is more complicated because it involves not only one latent GP, but one GP for each class. In the common multi-class likelihoods, as e.g. the softmax function, the GPs are coupled. This leads to complicated multivariate integrals 
which make a direct application of variational inference techniques intractable. Previous inference methods for the softmax model rely on approximations and do not scale \citep{Williams98bayesianclassification, DBLP:journals/jmlr/Chai12}.

To tackle this issue, \citet{hernandez2011robust} propose an alternative to the softmax, the {\it robust-max} likelihood. 
This likelihood simplifies the problem by focusing mainly on the maximal latent GP and discarding information of the other less likely classes.
The model is robust against outliers and often yields good classification accuracy. However, it sacrifices the gradual response of the traditional softmax for an all-or-nothing criterion 
leading to bad uncertainty quantification.

In problems with well separated classes and a few outliers, the robust-max likelihood is an excellent choice, while in problems with overlapping classes a gradual classification criterion is more desirable \citep{xiong2010classification}.
In this work, we introduce a novel likelihood, the {\it logistic softmax} likelihood, which combines the best of both worlds. It has a gradual classification criterion similar to the traditional softmax, but on the other hand also enables fast inference.

We propose an augmentation approach that renders the model conditionally conjugate. Inference in the augmented model is much easier. We derive a fast {\it variational inference} algorithm based on closed-form updates. Our inference approach is faster and more stable than the state of the art since it uses
efficient block coordinate ascent updates and does not rely on sampling.

Alternatively, the conditionally conjugate form of the augmented model directly leads to another inference strategy.
If we are willing to pay more computation time, we obtain {\it exact samples} from the true posterior by a Gibbs sampling scheme.
Our main contributions are as follows:
\begin{itemize}
	\item We introduce a new multi-class GP classification model building on a modification of the softmax likelihood function. By applying a variable augmentation approach, we render the model conditionally conjugate.
	\item We propose an efficient stochastic variational inference scheme which is based on block coordinate-ascent updates. Unlike in previous work, all updates are given in closed-form and do not rely on numerical quadrature methods or sampling.
	\item  Our method scales to datasets with many data points and a large number of classes. The experiments show that our method is faster than the state-of-the-art while leading to competitive prediction performance. 
	\item We solve the calibration issue of the robust-max likelihood as our model leads to much better uncertainty quantification.
\end{itemize}

The paper is structured as follows. Section~\ref{sec:related} introduces the problem of multi-class GP classification and reviews related work. In Section~\ref{sec:model} we introduce the new model and present a data augmentation strategy that renders the model conditionally conjugate.
In Section~\ref{sec:inference} we present an efficient inference algorithm.
We show experimental
results in Section~\ref{sec:experiments}.
Finally, Section~\ref{sec:conclusion} concludes and lays out future research
directions. Our code is included in a Julia package\footnote{\url{https://github.com/theogf/AugmentedGaussianProcesses.jl}}.

\section{Background and related work}
\label{sec:related}
We begin our review by introducing the multi-class GP classification model. Related work can be grouped into approaches that consider alternative likelihood functions or apply data augmentation strategies.

\paragraph{Multi-class GP classification.}
We consider a dataset of $N$ data points $\bX = (\bx_1,\dots,\bx_N)$ with labels $\by = (y_1,\dots,y_N)$, where $y_i \in \{1,\dots,C\}$ and $C$ is the total number of classes.
The multi-class GP classification model consists of a latent GP prior for each class $\boldf = (f^1, \dots, f^C)$, where $f^c \sim \GP(0, k^c)$ and $k^c$ is the corresponding kernel function. The labels are modeled by a categorical likelihood
\begin{equation}
p(y_i = k | \bx_i, \boldf_i) = g^k( \boldf(\bx_i)),
\label{eq:generalmodel}
\end{equation}
where $g^k(f)$ is a function that maps the real vector of the GP values $\bf$ to a probability vector.

The most common way to form a categorical likelihood is through the softmax transformation
\begin{align}
p(y_i = k | \boldf_i) = \frac{\exp\left(f^k_i\right)}{\sum_{c=1}^C \exp\left(f^c_i\right)},\label{eq:softmax}
\end{align}
where we use the shorthand $f^c_i = f^c(x_i)$ and for the sake of clarity we omit the conditioning on $x_i$.

There have been several early works addressing multi-class GP classification with a softmax likelihood
\citep{Williams98bayesianclassification, Kim:2006:BGP:1175897.1176234,
	DBLP:journals/jmlr/Chai12, Riihimaki:2013:NEP:2502581.2502584}.
Nevertheless, these methods do not scale well with the number of data points. \citet{DBLP:conf/aistats/IzmailovNK18} use tensor train decomposition to use high numbers of inducing points but do not provide efficient closed-form updates.

\paragraph{The robust-max likelihood.}
Recently, there have been advances to scale multi-class GP classification to big datasets by changing the likelihood.
\citet{hernandez2011robust} propose the \textit{robust-max} likelihood
\begin{align}
p(y=k|\boldf)=(1-\epsilon)\prod_{c\neq y}^C \Theta\left(f^k - f^c\right) + \frac{\epsilon}{C}, \label{eq:robustmax}
\end{align}where $\epsilon$ is the probability of a labeling error, and $\Theta$ is the Heaviside function. This likelihood simplifies the problem as it leads to a decoupling of the latent GPs.

Originally, the authors propose an expectation propagation (EP) based approach which only scales to small datasets. \citet{DBLP:conf/nips/HensmanMFG15} and \citet{DBLP:conf/aistats/SalimbeniEH18} scale this model to big datasets employing a variational inference approach but rely on numerical quadrature. As we show later, this likelihood has the big disadvantage of leading to poor confidence calibration.

\paragraph{The Heaviside likelihood.}
\citet{DBLP:conf/icml/Villacampa-Calvo17} build on the Heaviside likelihood
\begin{align}
p(y=k|\boldf)=\prod_{c\neq y}^C \Theta\left(f^k - f^c\right), \label{eq:robustmax2}
\end{align}
where $\Theta$ is again the Heaviside function.
The authors propose a scalable expectation propagation approach but have to make approximations on the likelihood. The inference is still slow and the applicability to big datasets is limited.

\paragraph{Data augmentation.}
Other approaches consider probabilistic data augmentation.
\citet{Wenzel_PG} propose an augmentation approach for binary GP classification leading to a conditionally conjugate model, but are limited to the binary classification setting.
\citet{DBLP:conf/nips/LindermanJA15} consider data augmentation for multinomial likelihoods but focus on sampling. The approach has the  disadvantage of breaking the symmetry between the classes and is limited to small datasets.
\citet{PG} propose conditionally conjugate P\'olya-Gamma augmentation for the softmax likelihood (extended by \citet{10.1371/journal.pone.0180343} to GPU support) which is suitable for sampling but cannot be used for obtaining an efficient variational inference algorithm since the ELBO is intractable.
\citet{girolami2006variational} propose an augmentation strategy to multinomial probit regression but does not scale.
\citet{DBLP:conf/icml/RuizTDB18} propose an augmentation approach for enabling subsampling of classes for parametric models with categorical likelihoods. The approach is limited to parametric models and cannot be applied to GP models.

\section{\methodname}
\label{sec:model}
We formulate a multi-class GP classification model which leads to well calibrated confidences and is amenable to fast inference.  
We define a new likelihood function, termed the {\it \namelikelihood~}, which shares the good prediction properties of the softmax. But in addition, it has the advantage that it allows for a data augmentation approach which renders the model conditionally conjugate. 
The augmented posterior can then be efficiently approximated by a structured mean-field variational inference method resulting in a fast algorithm with closed-form updates.


\subsection{The logistic-softmax GP model}
\label{sec:original_GP_model}
We consider the multi-class GP classification model as described in eq.~\ref{eq:generalmodel}.
Different functions $g$ for mapping real vectors to probability vectors that have been considered in literature include the softmax (eq.~\ref{eq:softmax}), the multinomial probit \citep{albert93}, the robust-max likelihood (eq.~\ref{eq:robustmax}) and the Heaviside likelihood (eq.~\ref{eq:robustmax2}).

In this work, we propose the {\it \namelikelihood~:}
\begin{equation}
p(y_i = k | \boldf_i)=\frac{\sigma\left(f^k_i\right)}{\sum_{c=1}^C \sigma\left(f^c_i\right)},
\label{eq:likelihood}
\end{equation}
where $\sigma(z)=(1+\exp(-z))^{-1}$ is the logistic function. 
Our likelihood is a modified version of the softmax likelihood which replaces the inner exponential functions by logistic functions. Alternatively, it can be interpreted as the standard softmax applied to a non-linearly transformed GP, i.e. $p(y_i | \boldf_i) = \text{softmax}( \log \sigma(\boldf_i))$. The likelihood reduces to the binary logistic likelihood for $C=2$.

In the following section we derive a three steps augmentation scheme, where we (i) decouple the GP latent variables $f^k_i$  in the denominator by the introduction of a set of auxiliary $\lambda$-variables, (ii) further simplify the model likelihood by introducing  Poisson random variables, and finally (iii) use a  P\'olya--Gamma representation of the sigmoid function~\citep{PG} to achieve the desired conjugate representation of the model.

\begin{figure}[H]
	\includegraphics[width=\columnwidth]{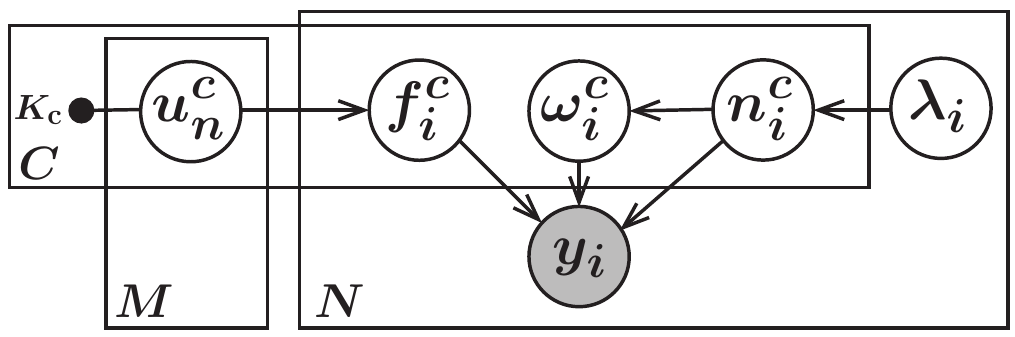}
	\caption{The final augmented model as presented in Section~\ref{sec:aug_model}. Shaded circles represent observable variables, empty circles latent variables and dots hyperparameters.}
	\label{fig:plate}
\end{figure}
\subsection{Towards a conjugate augmentation}
\label{sec:aug_model}

We expand the logistic-softmax  likelihood~\eqref{eq:likelihood} by three data augmentation steps leading to a conditionally conjugate model. 
The final model is displayed in Figure~\ref{fig:plate}. In the following we present the augmentations.

\paragraph{Augmentation 1: Gamma augmentation.}
To remedy the intractable normalizer term we make use of the integral identity
$\frac{1}{z}=\int_0^\infty \exp(-\lambda z)d\lambda$
and express the likelihood \eqref{eq:likelihood} as
\begin{align*}
p(y_i = k |\boldf_i) &= \frac{\sigma\left(f^k_i\right)}{\sum_{c=1}^C \sigma\left(f^c_i\right))}\\
&= \sigma\left(f^k_i\right) \int_0^\infty \exp\left(-\lambda_i \sum_{c=1}^C \sigma\left(f^c_i\right)\right) d\lambda_i.
\end{align*}

This augmentation is well known in the Gibbs sampling community to deal with intractable normalization constants (see e.g. ~\citet{walker2011posterior}) but is not often used in the setting of variational inference. By interpreting $\lambda_i$ as an additional latent variable we obtain the augmented likelihood
\begin{equation}\label{eq:afteraugmentation1}
p(y_i=k\vert \boldf_i,\lambda_i) = \sigma(f_i^k)\prod_{c=1}^C\exp\left(-\lambda_i\sigma(f_i^c)\right),
\end{equation}
and we impose the improper prior ${p(\lambda_i) \propto \mathds{1}_{[0,\infty)}(\lambda_i)}$. 
The improper prior is not problematic since it leads to a proper complete conditional distribution as we will see in the end of the section.

\paragraph{Augmentation 2: Poisson augmentation.}
We rewrite the exponential factors in~\eqref{eq:afteraugmentation1} based on the moment generation function of the Poisson distribution $\Po(\cdot\vert \lambda)$ which is
\begin{equation*}
\exp(\lambda(z - 1)) = \sum_{n=0}^\infty z^n \Po(z\vert \lambda).
\end{equation*}

Using $z=\sigma(-f)$ and the fact that $\sigma(f) = 1-\sigma(-f)$ we rewrite the exponential factors as
\begin{align*}
\exp\left(-\lambda_i\sigma(f_i^c)\right) & = \exp\left(\lambda_i(\sigma(-f_i^c) - 1)\right) \\
& = \sum_{n_i^c=0}^\infty(\sigma(-f_i^c))^{n_i^c}\Po(n_i^c\vert \lambda_i),
\end{align*}
which leads to the augmented likelihood
\begin{equation}\label{eq:afteraugmentation2}
p(y_i=k\vert \boldf_i,\lambda_i,\bn_i) = \sigma(f_i^k)\prod_{c=1}^C (\sigma(-f_i^c))^{n_i^c},
\end{equation}
where $\bn_i=(n_i^1,\ldots,n_i^C)$ and the augmented Poisson variables are distributed as
$p(n_i^c\vert \lambda_i)=\Po(n_i^c\vert \lambda_i)$, see e.g. \cite{Donner2017, Donner2018a}.
Note that this augmentation is only possible since the transformation on $f_i^c$ is bounded, hence the need for a modified likelihood.

\paragraph{Augmentation 3: P\'olya-Gamma augmentation.} 
In the last augmentation step, we aim for a Gaussian representation of the sigmoid function. The P\'olya-Gamma representation~\citep{PG} allows for rewriting the sigmoid function as a scale mixture of Gaussians
\begin{equation}\label{eq:pgtransform}
\sigma(z)^n = \int_{0}^\infty 2^{-n}\exp\left(\frac{nz}{2} - \frac{z^2}{2}\omega\right)\PG(\omega\vert n, 0),
\end{equation}
where $\PG(\omega\vert n,b)$ is a P\'olya-Gamma distribution. P\'olya-Gamma variables are well suited for augmentations since the moments are known analytically and an efficient sampler exists~\citep{PG}.
By applying this augmentation to~\eqref{eq:afteraugmentation2} we obtain
\begin{equation}\label{eq:afteraugmentation3}
\begin{split}
&p(y_i=k\vert \boldf_i,\lambda_i,\bn_i,\bomega_i) =\\ &\prod_{c=1}^C 2^{-(y'^c_i+n_i^c)}\exp\left(\frac{(y'^c_i-n_i^c)f_i^c}{2}-\frac{(f_i^c)^2}{2}\omega_i^c\right),
\end{split}
\end{equation}
where $\bomega_i=(\omega_i^1,\ldots,\omega_i^C)$ are P\'olya-Gamma variables with distributions
\begin{align*}
p(\bomega_i | \bn_i,y_i) =\prod_{c=1}^C\PG(\omega_i^c\vert y'^c_i + n_i^c, 0),
\end{align*}
where $\by'$ is an $N\times C$-dimensional one-hot encoding of the labels
, i.e. $y'^c_i $ is $1$ if $y_i=c$, and $0$ otherwise.
Details are deferred to appendix~\ref{app:pg}.

Realizing that~\eqref{eq:afteraugmentation3} has a Gaussian form with respect to $\boldf_i$ we achieved our goal of a conjugate representation of the latent GPs. As we will show in the next paragraph the model is also conditionally conjugate for the augmented variables.


\paragraph{The final model.}
The effort of the augmentations finally pays off as the final augmented model is now tractable and the complete conditional distributions are given in closed-form.


The complete conditionals of the GPs $\boldf^c$ are
\begin{equation*}
p(\boldf^c \mid \by, \bomega^c, \bn^c) = \mathcal{N}\left(\boldf^c \mid \frac{1}{2}A^c(\by'^c-\bn^c), A^c\right),
\end{equation*}
where the conditional covariance matrix is given by $A^c = \left(\diag(\bomega^c)+K_c^{-1}\right)^{-1}$ and $K_c$ is the kernel matrix of the GP $\boldf^c$.
For the conditional distribution of $\blambda$ we get
\begin{equation*}
p(\lambda_i\mid \bn_i) = \Ga\left(\lambda_i\mid 1+\sum_{c=1}^C n_i^c,C\right),
\end{equation*}
where $\Ga(\cdot\vert a,b)$ denotes a gamma distribution with shape parameter $a$ and rate parameters $b$. The improper prior on $\lambda_i$ does not impose an issue since the complete conditional distribution is proper.

For the Poisson variables $\bn$, we get
\begin{equation*}
p(n_i^c\mid f_i^c,\lambda_i) = \Po \left(n_i^c\mid\lambda_i \sigma(f_i^c)\right),
\end{equation*}
Finally, for the P\'olya-Gamma variables $\bomega$ the complete conditional distributions are
\begin{align*}
p(\omega_i^c\mid  n_i^c, f_i^c,y_i) &= \PG \left(\omega_i^c\mid {y'}_i^c + n_i^c,|f_i^c|\right).
\end{align*}

\section{Inference} \label{sec:inference}
We derive a variational approximation of the posterior of the augmented model \eqref{eq:afteraugmentation3}. In the following we develop an efficient stochastic variational inference (SVI) algorithm that is based on closed-form block coordinate ascent updates.
Our method allows both for subsampling of data points and of outcomes (classes) scaling to datasets with a large number of data points and a large number of classes.

\subsection{Variational approximation}
\label{sec:var_approximation}
To scale our model to big datasets, we approximate the latent GPs $\boldf^c$ by {\it sparse GPs} building on {\it inducing points}.
For each GP $\boldf^c$, we introduce $M$ inducing points $\bu^c$ and connect the GP values with the inducing points via the joint prior distribution $p(\boldf^c, \bu^c)$ given in \citet{Titsias09variationallearning}. Details on variational sparse GP approximations can be found in \citet{Titsias09variationallearning, hensman2013gaussian}.

We approximate the posterior distribution of the latent sparse GPs $\bu$ and the augmented variables $\blambda,\bn, \bomega$ by
assuming the following structure of the variational distribution
$q(\bu, \blambda,\bn, \bomega) = q(\bu, \blambda)q(\bn, \bomega)$.
Note that the only assumption on the variational posterior is the decoupling of two groups of variables. \textbf{}
Since our model is conditionally conjugate, the family of the optimal variational distribution can be easily determined 
by averaging the complete conditionals in log-space~\citep{DBLP:journals/corr/BleiKM16}.
From the above decoupling assumption, it follows that the optimal variational posterior has a factorizing form $q(\bu, \blambda,\bn, \bomega) =q(\bu)q(\blambda)q(\bomega,\bn)$ and the factors are
\begin{alignat*}{2}
q(\bu)&= \prod_c\mathcal{N}(\bu^c|\bmu^c,\Sigma^c),
\; q(\blambda)= \prod_i \Ga(\lambda_i|\alpha_i,\beta_i),\\
q(\bomega,\bn)&= \prod_{i,c} \PG(\omega_i^c|{y'}_i^c+n_i^c,b_i^c)\Po(n_i^c|\gamma_i^c),
\end{alignat*}
where $\bmu^c$, $\Sigma^c$, $\alpha_i$, $\beta_i$, $b_i^c$, $\gamma_i^c$, for all $i \in \{1,\dots,N\}$ and $c \in \{1,\dots,C\}$ are the {\it variational parameters}. 
The variational parameters are optimized by a coordinate ascent scheme outlined in Section~\ref{sec:SVI}.
Finally, the approximate posterior of the sparse GPs $q^*(\bu)$ can be used to obtain an approximate posterior of the original latent GPs $\boldf$ by $q^*(\boldf):=\int p(\boldf|\bu)q(\bu) d\bu$ which is given in closed-form \citep[see e.g.,][]{Hensman2015}. 


\subsection{Stochastic variational inference}
\label{sec:SVI}
Building on the conditionally conjugate representation of our model deriving efficient variational parameter updates is straightforward. We implement the classic SVI algorithm described by \citet{JMLR:v14:hoffman13a}, which builds on block coordinate ascent updates. We iteratively optimize each factor of the variational distribution, while holding the others fixed. The variational parameters of each factor are directly set to the optimal value given the other parameters.

We compute the block coordinate ascent (CAVI) updates in closed-form by averaging the parameters of each complete conditional in log space \citep{DBLP:journals/corr/BleiKM16} and details are deferred to appendix~\ref{appendix:updates}.
When using minibatches of the data, each global variational parameter (i.e. $\bmu^c$ and $\Sigma^c$) is updated using a convex combination of the old parameter and the CAVI update, which corresponds to a natural gradient ascent scheme \citep{JMLR:v14:hoffman13a}. 
Remarkably, the negative ELBO in our augmented model is convex in the global parameters (see appendix~\ref{appendix:convex} for the proof). Therefore, our algorithm is ensured to converge to the global optimum \citep{JMLR:v14:hoffman13a}.
The inference algorithm is summarized in Alg.~\ref{algo} and its complexity is $\mathcal{O}(CM^3)$. 

\begin{algorithm}[!ht]
	\caption{\methodname}
	\begin{algorithmic}[1]
		\State \textbf{Input:} data $\bX$,$\by$, minibatch size $|\mathcal S|$
		\State \textbf{Output:} variational posterior GPs $p(u^c | \mu^c, \Sigma^c)$
		\State Set the learning rate schedules $\rho_t, \rho^h_t$ appropriately
		\State Initialize all variational parameters and hyperparameters
		\State Select $M$ inducing points locations (e.g. kMeans)
		\For {iteration $t=1,2,\dots$}
		\State {\tt \# Sample minibatch:}
		\State Sample a minibatch of the data ${\mathcal S} \subset \{1,\dots,N\}$
		
		\State {\tt \# Local variational updates}
		\For {$i \in \mathcal S$}
		\State Update $(\alpha_i, \bgamma_i)$ \label{algo:line_alpha_lambda} (Eq.~\ref{eq:local_updates_gamma},\ref{eq:local_updates_alpha})
		\For {each class $c$}
		\State Update $b_i^c$  (Eq.~\ref{eq:local_updates_omega}) \label{algo:line_local}
		\EndFor
		\EndFor
		
		\State {\tt \# Global variational GP updates}
		\For {each class $c$}
		\State $\mu^c \gets (1-\rho_t) \mu^c + \rho_t \hat \mu^c $ (Eq.~\ref{eq:global_updates_1}) \label{algo:line_global1}
		\State $\Sigma^c \gets (1-\rho_t) \Sigma^c + \rho_t \hat \Sigma^c $ (Eq.~\ref{eq:global_updates_2}) \label{algo:line_global2}
		\EndFor
		\State {\tt \# Hyperparameter updates}
		\State Gradient step $h \gets h + \rho^h_t \nabla_h \mathcal L$ \label{algo:line_hyper}
		\EndFor
		
	\end{algorithmic}
	\label{algo}
\end{algorithm}

\paragraph{Extreme classification.}
When the number of possible outcomes (classes) $C$ is very large, using probabilistic multi-class models becomes generally computationally expensive as the likelihood (categorical distribution) scales linearly with the number of classes.
Using large categorical distributions is a challenging problem \citep{DBLP:conf/icml/RuizTDB18,aueb2016one}.

With a slight modification, our method can deal with an extreme classification setting (large number of classes). In our augmentation, the GPs in the normalizer term are decoupled and allow for subsampling of the classes. This reduces the complexity to $\mathcal O(M^3)$, i.e. being independent of the number of classes.
We provide details in appendix~\ref{appendix:subclasses}.
This approach is especially useful when using shared hyperparameters among the class specific latent GPs.


\paragraph{Predictions.}
The posterior distribution of the latent function $p(f_\star^c|x_\star,\by)$ at a new test point $x_\star$ is approximated by
\begin{equation*}
q(f_\star^c|x^\star,\by) = \int p(f_\star^c|\bu^c) q(\bu^c)d\bu=\mathcal{N}\left(f_\star^c|{\mu_\star}^c,{\sigma_\star^2}^c\right),
\end{equation*} where the mean is $\mu_\star^c={K_{\star m}}^c{K_{mm}^{-1}}^c\bmu^c$ and the variance ${\sigma^2_\star}^c={K_{\star\star}}^c + {K_{\star m}}^c{K_{mm}^{-1}}^c(\Sigma^c {K_{mm}^{-1}} ^c - I){K_{m\star}}^c$.  The matrix $K_{\star m}$ denotes the kernel matrix between the test point and the inducing points and $K_{\star\star}$ the kernel value of the test point.
The final approximate predictive distribution of a test label is
\begin{align*}
p(y=k|x_\star,\by) &\approx \int p(y=k|\boldf_\star)\prod_{c=1}^C q(f_\star^c|x^\star,\by)d\boldf^\star,
\end{align*}
where $p(y=k|\boldf_\star)$ is the logistic-softmax likelihood.
This is a $C$-dimensional analytically intractable integral. We approximate it by Monte Carlo integration. 
For faster convergence, the random samples can be replaced by Quasi-Monte Carlo sequences \citep{Owen1998, buchholz2018quasi}.
Finally, a point is classified by the highest predictive likelihood, ${y^*_i} =\arg \max_{c\in C} p\left(y_i=c\mid\boldf\right)$.

\paragraph{Optimization of the hyperparameters.}
We select the optimal kernel hyperparameters by maximizing the marginal likelihood $p(y|h)$, where $h$ denotes the set of hyperparameters
(this approach is called empirical Bayes \citep{EB89}).
We follow an approximate approach and optimize the fitted variational lower bound ${\cal L}(h)$ 
as a function of $h$ by alternating between optimization steps w.r.t. the variational parameters and the hyperparameters \citep{cSGD}.

\subsection{Gibbs sampling}
\label{sec:Gibbs}
Since our augmented model is conditionally conjugate we can directly derive a Gibbs sampling scheme. In order to sample from the {\it exact posterior}, we alternate between drawing a sample from each complete conditional distributions. The augmented variables are naturally marginalized out and asymptotically, the latent GP samples will be from the true posterior.

\section{Experiments}
\label{sec:experiments}
\begin{figure*}
	\centering
	\includegraphics[width=\textwidth]{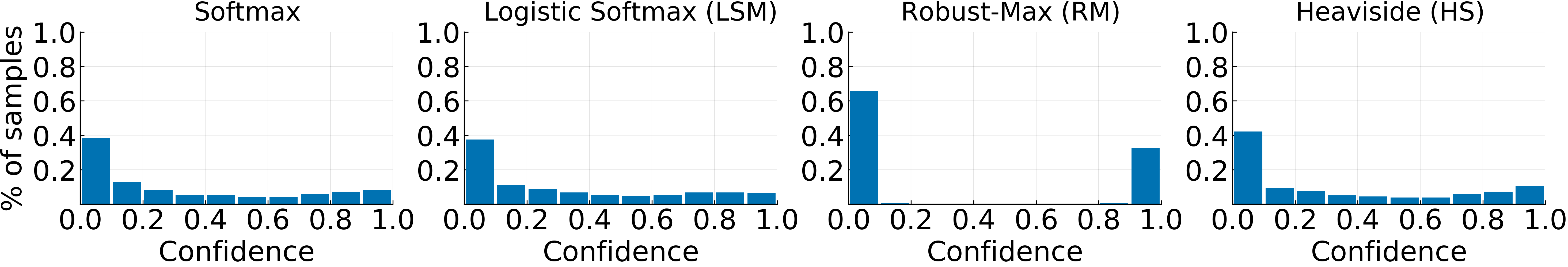}
	\includegraphics[width=\textwidth]{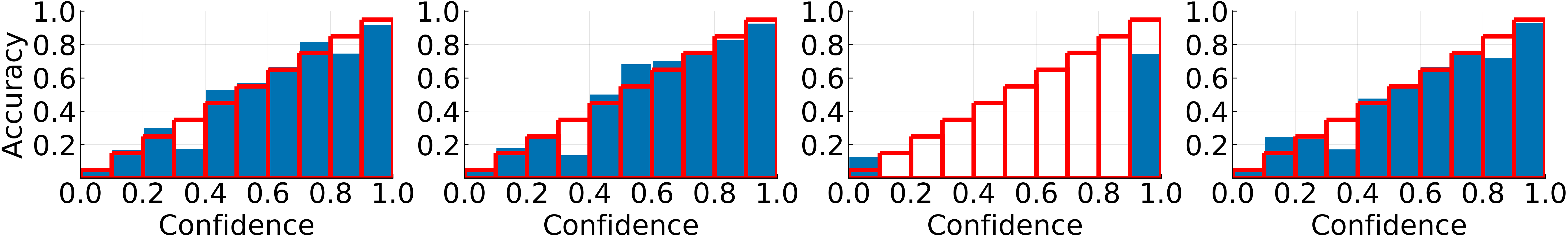}
	\caption{
		{\it Likelihood comparison:} Confidence histograms (top) and reliability diagrams (bottom) for four different likelihood models. The robust-max model always predicts with probability either close to one or close to zero leading to a poor confidence  calibration.
	}
	\label{fig:bins_likelihood}
\end{figure*}

In this section we empirically answer the following questions:
\begin{itemize}
	\item What is the effect of using the softmax, logistic-softmax, robust-max and Heaviside likelihood on predictive performance and calibration quality? (Section~\ref{sec:exp_comp})
	\item How does the augmentation affect the predictive performance? (Section~\ref{sec:exp_augm})
	\item How does our method perform compared to other state-of-the-art GP based multi-class classification methods? (Section~\ref{sec:exp_numerical_comp})
\end{itemize}
In all experiments we use a squared exponential covariance function with automatic relevance determination (ARD): $k(\boldsymbol{x},\boldsymbol{x'})=\eta\exp\left(-\sum_{d=1}^D \frac{\left(x_d - x_d'\right)^2 }{2l_d^2}\right)$, where we set the initial
variance $\eta$ to 1 and the length scales $\boldsymbol{l}$ are initialized to the median of the pairwise distance matrix of the data. The hyperparameters are optimized using Adam \citep{kingma2014adam}.
We use a collection of datasets from the LIBSVM repository\footnote{\url{https://www.csie.ntu.edu.tw/\~cjlin/libsvmtools/datasets/multiclass.html}}.
Every dataset has been normalized to mean 0 and variance 1. 
For each method, we use 200 inducing points, unless stated otherwise. The initial inducing points locations are determined by the kmeans++ algorithm \citep{arthur2007k}. We find that fixing the locations while training gives good results. We use a mini-batch size of 200 and all experiments are performed on a single CPU.

\subsection{Comparison of the different likelihoods}

\label{sec:exp_comp}
We begin the experiments by investigating the effect of using different likelihood functions. We compare
our novel logistic-softmax (eq.~\ref{eq:likelihood}), the softmax (eq.~\ref{eq:softmax}), the robust-max (eq.~\ref{eq:robustmax}) and the Heaviside likelihood (eq.~\ref{eq:robustmax2}). For each model we employ variational inference to obtain an approximate posterior. In this experiment, no augmentation is used and the gradients are estimated by sampling.

To investigate uncertainty calibration, we create seven different toy datasets of 500 points with three classes. The data is generated from a mixture of Gaussians model with different variances $\sigma^2$. For $\sigma^2=0$, the classes are sharply separated and for $\sigma^2=1$, the classes highly overlap and are almost indistinguishable.

\begin{figure}[h]
	\centering
	\includegraphics[width=\columnwidth]{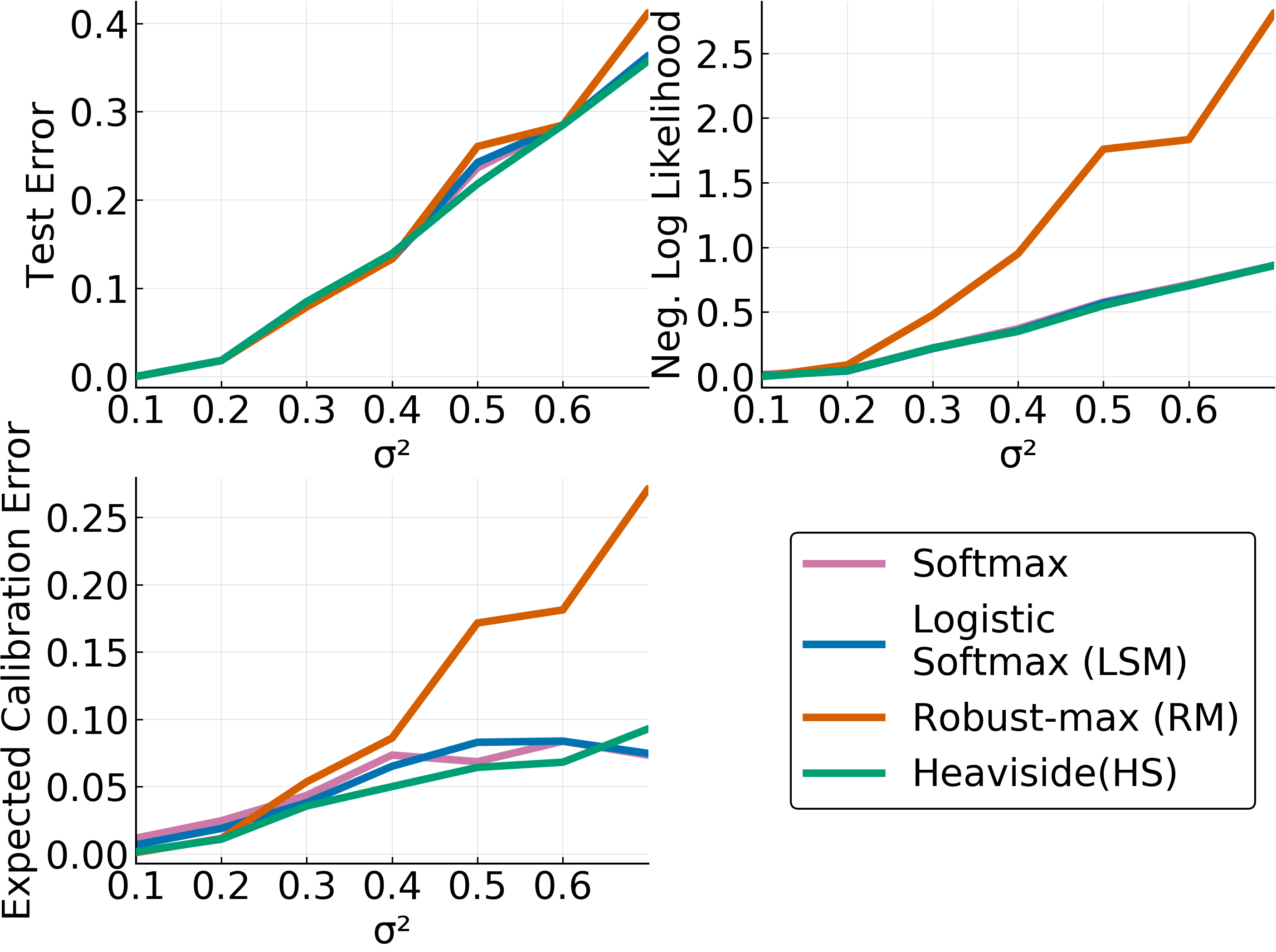}
	\caption{{\it Likelihood comparison:} The test error, negative log-likelihood and calibration error are plotted as function of the noise ($\sigma^2$) in the generated dataset. 
		For highly overlapping classes (large $\sigma^2$), the robust-max likelihood yields poor calibration and bad log-likelihood values.}
	\label{fig:comp_likelihood}
\end{figure}

See appendix~\ref{appendix:decision_boundary} for a visualization of the decision boundaries of the different methods. In Figure~\ref{fig:comp_likelihood} we plot test error, negative log-likelihood and calibration error as function of the noise in the data. 
The (expected) calibration error is a summary statistic of calibration and is computed by the expectation between confidence and accuracy in the reliability diagram \citep[c.f.][]{DBLP:conf/icml/GuoPSW17}.

For datasets where the classes are sharply separated (small $\sigma^2$), all models perform similarly.
But for datasets where classes overlap (high $\sigma^2$), the robust-max performs poorly due to bad uncertainty calibration.

In Figure~\ref{fig:bins_likelihood} we show the confidence histograms and reliability diagrams for one dataset ($\sigma^2=0.5$). 
The diagrams are generated according to \citet{naeini2015obtaining,DBLP:conf/icml/GuoPSW17} -- the reliability diagram displays the accuracy as function of confidence (a perfectly calibrated model would produce the identity function) and the confidence histogram shows the empirical distribution of the prediction confidence.

The robust-max model fails to provide sensitive uncertainty estimates and only predicts with either probability close to zero or close to one.
The softmax, logistic-softmax and Heaviside likelihood yield similar predictive performance and confidence calibration. However, as the following experiments show, our approach is much faster than the softmax and Heaviside model. It is the only scalable approach that leads to well calibrated confidences and the logistic-softmax can be used as an efficient replacement of the standard softmax.

\subsection{Effect of the augmentation}
\label{sec:exp_augm}
\begin{figure}
	\includegraphics[width=\columnwidth]{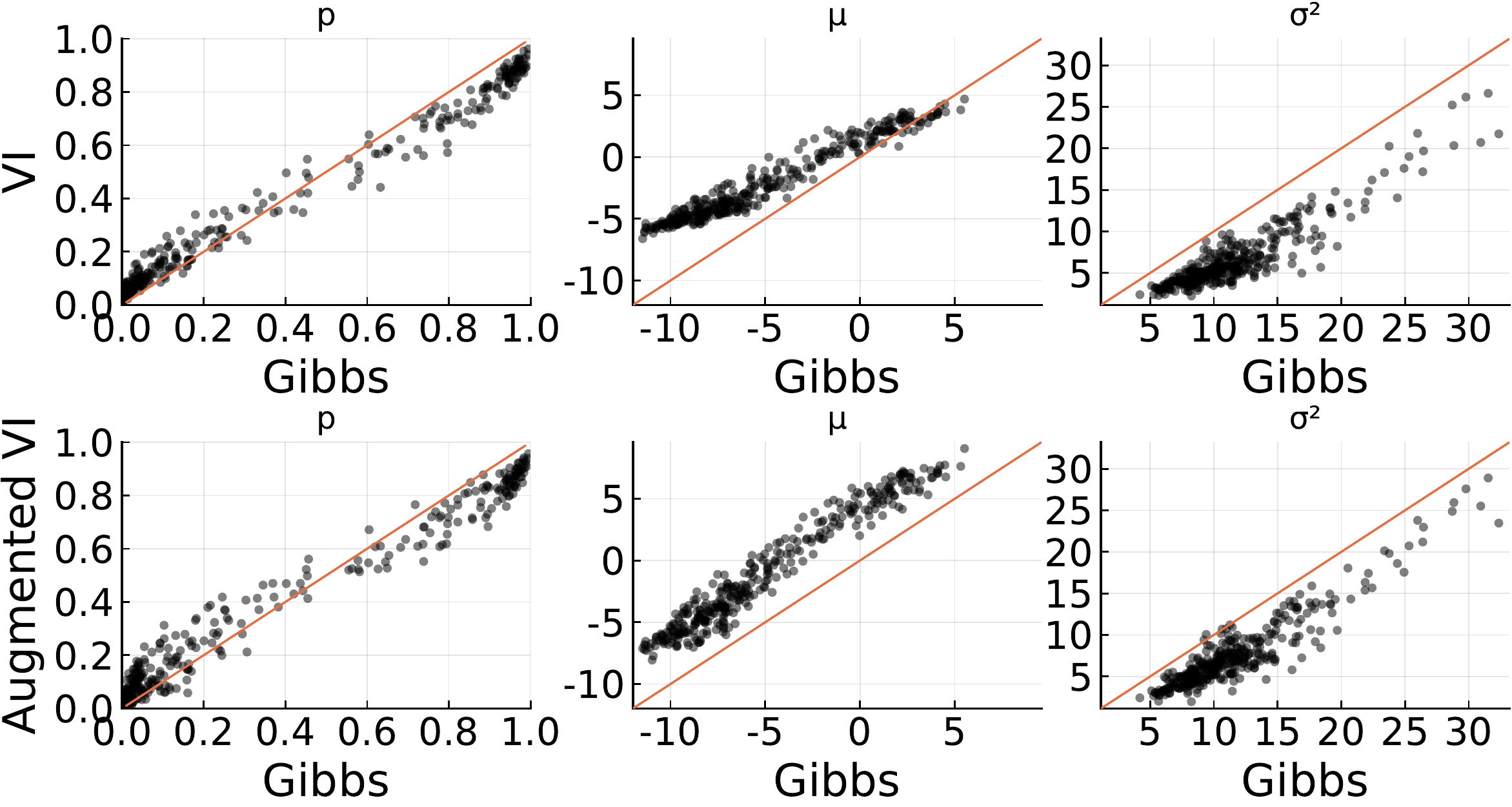}
	\caption{{\it Effect of the augmentation:}
		Comparison of the predictive marginals ($p$), posterior mean ($\mu$) and posterior variance ($\sigma^2$) on a test set. 
		Each plot shows the ground truth of the Gibbs sampler on the x-axis. On the y-axis the estimated values by variational inference without augmentation 
		\textit{VI} (top) and augmented variational inference \textit{Augmented VI} are shown (bottom). Our efficient augmented VI method produces values very close to the less efficient VI method. Both methods slightly overestimate the mean ($\mu$) and underestimate the variance ($\sigma^2$). However, for both methods the final predictions ($p$) are close to the ground truth.
	}
	\label{fig:inferencecomparison}
\end{figure}

We investigate the effect of the augmentation of the logistic-softmax model and its variational approximation. To this end we compare
three different inference methods (1) variational inference for our augmented model (\textit{Augmented VI}), (2) variational inference without augmentation (approximating the posterior of the original model from section~\ref{sec:original_GP_model} using a variational Gaussian), where the gradients are computed via sampling (\textit{VI}) and (3) Gibbs sampling (\textit{Gibbs}), c.f. Section~\ref{sec:Gibbs}.
After burn-in, the samples
from the Gibbs sampler serve as ground truth since they come from the exact posterior.
In this experiment we do not use the inducing point approximation and all hyperparameters are fixed. We apply all three methods on the dataset Wine (3 classes) and
compare the predictive likelihood ($p$) and the mean ($\mu$) and variance ($\sigma^2$) of the latent GPs on a test set. We compare each entry of the three-dimensional vectors $p$, $\mu$, $\sigma^2$ with the ground truth and display the
results for all classes $c=1,2,3$ combined in Figure~\ref{fig:inferencecomparison}.

Variational inference in the augmented model results in an approximate posterior which is very close to the variational inference solution in the original model. Both methods lead to a similar slight approximation error of the posterior mean $\mu$ and variance $\sigma^2$ and give predictive marginals $p$ close to the ground truth. The Gibbs sampling approach has a final prediction accuracy of 0.98, whereby both variational inference methods have a final accuracy of 0.96.
We find that the augmentation approach can be used as a scalable alternative to standard variational inference.

\subsection{Inducing points and hyperparameters}
\label{sec:exp_inducing}
In this experiment we answer two questions.
What is the effect of the number of inducing points and what is the difference between using shared hyperparameters and individual hyperparameters for each latent GP?
We train our model on the Shuttle dataset (58,000 points, 9 classes) for 200 epochs. We vary the number of inducing points from 5 to 400, and set the GP hyperparameters to be either shared or independent among classes. 

\begin{figure}[!ht]
	\centering
	\includegraphics[width=\columnwidth]{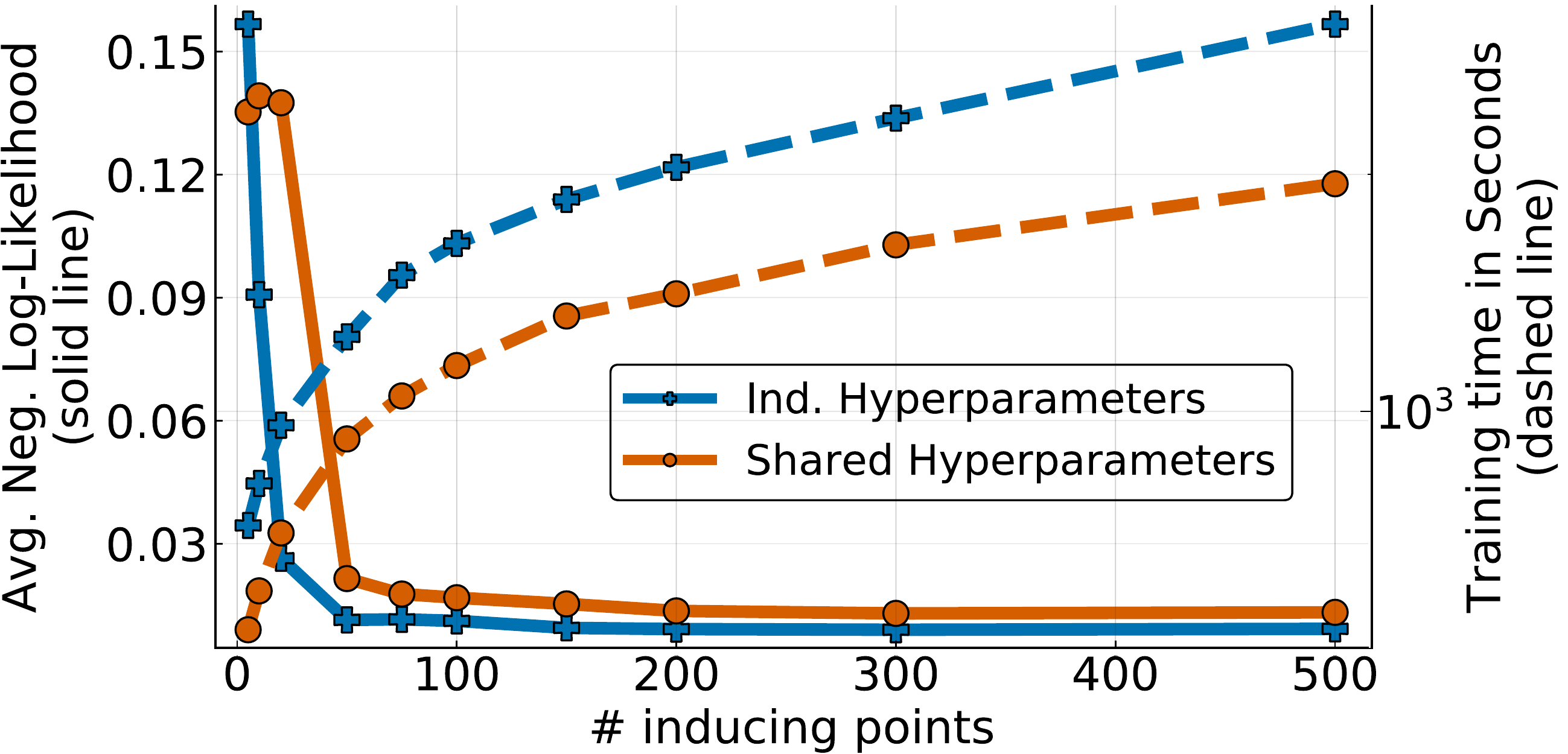}
	\caption{{\it Inducing points and hyperparameters:} The trade-off between predictive performance and run time is shown. Two versions of our method are used: individual hyperparameters for each GP (blue) and shared hyperparameters (orange). On the left y-axis we plot the negative log-likelihood (solid line) and on the right y-axis the training time (dashed line) as function of the number of inducing points.}
	\label{fig:indpoints}
\end{figure}

In Figure~\ref{fig:indpoints} we display the trade-off between predictive performance and training time.
We plot the  negative log-likelihood (solid lines, y-axis left) and training time (dashed lines, y-axis right) as a function of the number of inducing points.
If the number of inducing points is increased, the negative log-likelihood goes down and, oppositely, the training time goes up. We find that using only 200 inducing points already leads to near optimal predictive performance. 
Using independent hyperparameters over shared hyperparameters does not lead to a significant improvement of the predictive performance but implies a higher computational cost, especially for datasets with a large number of classes.

\begin{figure*}
	\centering
	\includegraphics[width=0.95\textwidth]{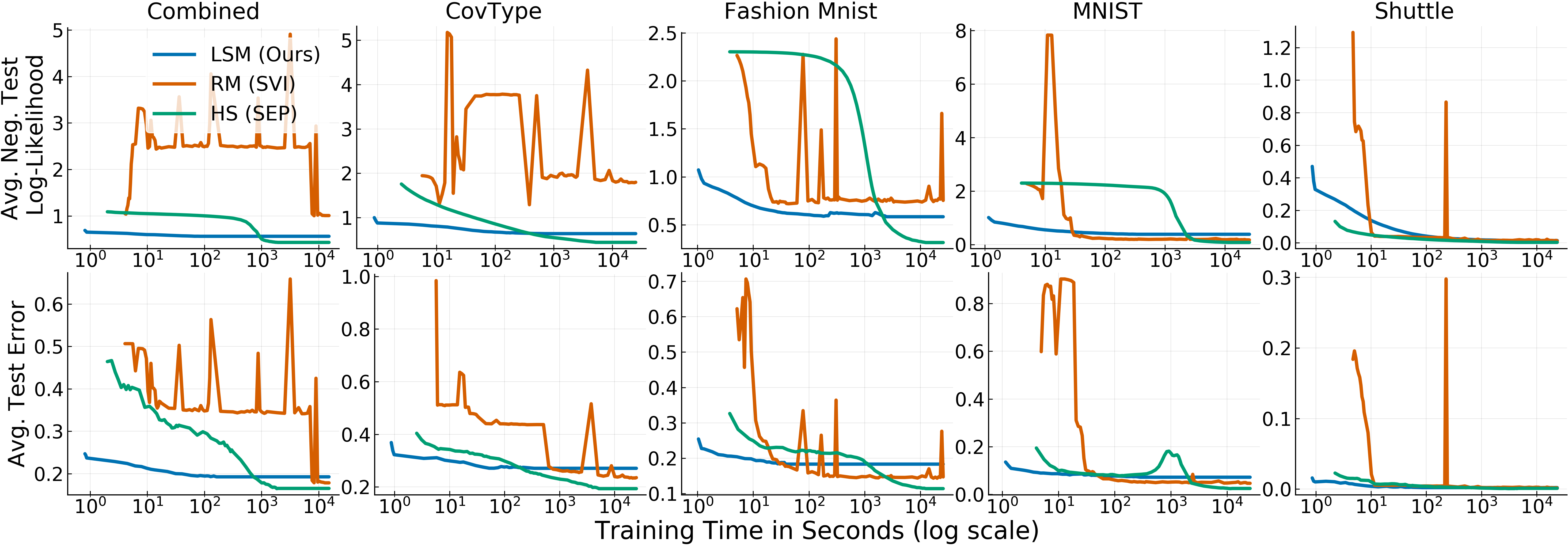}
	\caption{{\it Numerical comparison:} Prediction error and negative log-likelihood as a function of training time (seconds on a $\log_{10}$ scale).
		Our method (\LSM) converges one to two orders of magnitudes faster than the Heaviside model (\HS) and is around 10 times faster than the robust-max model (\RM). \textsc{rm} yields poor negative log-likelihood values due to poor uncertainty calibration.
	}
	\label{fig:conv}
\end{figure*}

\subsection {Numerical comparison}
\label{sec:exp_numerical_comp}
Finally, we evaluate the predictive performance and convergence speed of our method against other state-of-the-art multi-class GP classification approaches.
We compare our logistic-softmax likelihood based approach (\textsc{lsm}) against two competitors. First, the robust-max likelihood model (\textsc{rm}) by \citet{Hensman2015} which is provided in the package GPFlow
\citep{DeG.Matthews:2017:GGP:3122009.3122049} and trained by the natural gradient method of \citet{DBLP:conf/aistats/SalimbeniEH18} and second, the Heaviside likelihood model (\textsc{hs}) trained by a scalable EP method  \citep{DBLP:conf/icml/Villacampa-Calvo17}.
For all methods, the hyperparameters are initialized to the same values, and are optimized using Adam.
We compare the methods on five different multi-class benchmark datasets:
Combined (98,528 points, 50 features, 3 classes),
CovType (581,000 points, 54 features, 7 classes),
Fashion-MNIST (70,000 points, 784 features, 10 classes), MNIST (70,000 points, 784 features, 10 classes) and Shuttle (58,000 points, 9 features, 7 classes).

In Figure~\ref{fig:conv} we plot the test error and negative log-likelihood as functions of the training time for each dataset. 
We find that our method (\textsc{lsm}) is one to two orders of magnitude faster than the EP based method for the Heaviside model (\textsc{hs}) and around ten times faster than the SVI based method for the robust-max model (\textsc{rm}).

Furthermore, our method consistently beats \textsc{rm} in terms of negative log-likelihood due to the better calibrated uncertainty quantification. Only on the MNIST dataset \textsc{rm} reaches a slightly better log-likelihood. This dataset is easily separable and therefore, suits well to the robust-max likelihood assumptions. 
On most datasets, the EP based method (\textsc{hs}) leads to slightly better  predictive log-likelihood values, but is demanding a much longer training time.
In contrast to the log-likelihood, the pure prediction error is not very sensitive to uncertainty calibration. All three methods achieve similar prediction errors whereby \textsc{hs} is a bit better on some datasets.

Moreover, the optimization curves in Figure~\ref{fig:conv} show that our inference method is much more stable than the SVI approach for the \textsc{rm} model. This is due to our efficient coordinate ascent updates which are given in closed-form. The \textsc{rm} approach suffers from additional noise injected by approximating its gradients.

To summarize, our method is a good choice for fast inference on big datasets. 
It is particularly well fitted for datasets with overlapping classes where well calibrated uncertainty quantification is important. Due to the closed-form updates our method is more stable than the competitors.

\section{Conclusion}
\label{sec:conclusion}
We proposed an efficient Gaussian process multi-class classification method that builds on data augmentation. The augmented model is conditionally conjugate allowing for fast and stable variational inference based on closed-form updates.
The experiments show that our approach leads to better confidence calibration than recent scalable multi-class GP classification methods. Additionally, we achieve competitive prediction performance while being faster than state-of-the-art. For small problems the proposed Gibbs sampler can be used which provides samples from the exact posterior. 

The presented work shows how data augmentation can speed up inference in GP based models.
Our approach may pave the way to similar augmentation strategies for other Bayesian models.
Future work may aim at extending our approach to Bayesian neural networks (BNNs). Inference in BNNs is a hard problem. Exchanging the common softmax link functions with our proposed logistic-softmax may leads to a conditionally conjugate augmentation approach for BNNs.
Typically, Gaussian priors are used for the weights of the network. In the augmented model the posterior of the weights would be given in closed-form.
This might lead to an efficient inference algorithm.


\newpage
\subsubsection*{Acknowledgements}
We thank Stephan Mandt, Robert Bamler and Marius Kloft for discussions and feedback on the manuscript.
We also thank Simon Danisch for helping with implementation details in Julia.
This work was partly funded by the German Research Foundation (DFG) awards KL 2698/2-1 and GRK1589/2 and the
by the Federal Ministry of Science and Education (BMBF)
awards 031L0023A, 01IS18051A.

%

\bibliographystyle{apa}
\bibliography{bib}

\clearpage
\appendix
\newpage
\section{Appendix}

\subsection{Reparametrization of the P\'olya-Gamma variables}
\label{app:pg}
By applying the augmentation of the sigmoid~\eqref{eq:pgtransform} to the augmented likelihood~\eqref{eq:afteraugmentation2}, we obtain the P\'olya-Gamma augmented likelihood
\begin{align}
&p(y_i=k\vert \boldf_i,\lambda_i,\bn_i,\tilde{\omega}_i,\bomega_i) =\frac{1}{2}\exp\left(\frac{f^k_i}{2}-\frac{(f^k_i)^2}{2}\tilde{\omega}_i\right)\nonumber\\
&\times\prod_{c=1}^C 2^{-n_i^c}\exp\left(-\frac{n_i^cf_i^c}{2}-\frac{(f_i^c)^2}{2}\omega_i^c\right), \label{eq:intpg}
\end{align}
where we impose the prior distributions
\begin{align*}
p(\tilde{\omega}_i) =& \PG(1,0)\\
p(\bomega_i|\bn_i) =& \prod_c^C \PG(\omega_i^c|n_i^c,0).
\end{align*}
We simplify this expression by combining all terms corresponding to the index $k$.
To this end, we use a one hot-encoding of $\by \in \{0,\ldots,C\}^N$  as $\by' \in \{0,1\}^{C\times N}$,
\begin{equation*}
{y'}_i^c =  \left\{\begin{array}{c} 1 \text{ for } y_i = c\\ 0 \text{ otherwise.} \end{array} \right. .
\end{equation*} 
Building on the identity $\omega_1 + \omega_2 = \omega_3$ with $\omega_1 \sim \PG(b_1,c)$, $\omega_2 \sim \PG(b_2,c)$ and $\omega_3 \sim \PG(b_1+b_2,c)$,
we rewrite equation (\ref{eq:intpg}) as
\begin{equation*}\label{eq:afteraugmentation3_v2}
\begin{split}
&p(y_i=k\vert \boldf_i,\lambda_i,\bn_i,\bomega_i) =\\
&\prod_{c=1}^C 2^{-({y'}^c_i+n_i^c)}\exp\left(\frac{({y'}_i^c-n_i^c)f_i^c}{2}-\frac{(f_i^c)^2}{2}\omega_i^c\right),
\end{split}
\end{equation*}
where the terms corresponding to $\tilde{\bomega}$ are now absorbed into the terms corresponding to $\bomega$.

\subsection{Block coordinate ascent (CAVI) updates}
\label{appendix:updates}
The variational distribution is $q(\bu, \blambda, \bn, \bomega) =q(\bu)q(\blambda)q(\bomega,\bn)$ and the factors are
\begin{alignat*}{2}
q(\bu)&= \prod_c^C\mathcal{N}(\bu^c|\bmu^c,\Sigma^c),
\; q(\blambda)= \prod_i \Ga(\lambda_i|\alpha_i,\beta_i),\\
q(\bomega,\bn)&= \prod_{i,c} \PG(\omega_i^c|{y'}^c_i+n_i^c,b_i^c)\Po(n_i^c|\gamma_i^c).
\end{alignat*}
In the CAVI scheme \citep{JMLR:v14:hoffman13a} each factor is iteratively updated by the following equation. Suppose we want to update the variational distribution corresponding to the latent variable $\boldsymbol{\theta} \in \{\bu, \blambda, \bn, \bomega\}$. Let $\overline{\boldsymbol{\theta}}$ be the set of the other latent variables, then $q^*(\boldsymbol{\theta})$ is updated by
\begin{align}
q^*(\boldsymbol{\theta})\propto \exp \left(\expec{q(\overline{\boldsymbol{\theta}})}{\log p(\theta\mid\overline{\boldsymbol{\theta}})}\right).
\end{align}
Using this equation gives the closed-form update for each variational parameter.
\begin{align}
\begin{split}\overline{f_i^c}=&\sqrt{\mathbb{E}_{q(f^c)}\left[\left(f_i^c\right)^2\right] }\\
=& \sqrt{\widetilde{K}_{ii}^c+\kappa^c_i\Sigma^{c}{\kappa^c_i}^\top +  (\kappa^c_i\bmu^c)^\top \kappa^c_i\bmu^c}
\end{split}\nonumber\\
\gamma_i^c =& \frac{\exp\left(\psi(\alpha_i)\right)\exp\left(-\frac{\kappa_i^c\bmu^c}{2}\right)}{\beta_i\cosh\left(\frac{\overline{f_i^c}}{2}\right)} \label{eq:local_updates_gamma}\\
\alpha_i =& 1 + \sum_{c=1}^C \gamma_i^c, \quad \beta_i = C \label{eq:local_updates_alpha}\\
b_i^c =& \overline{f_i^c},\label{eq:local_updates_omega}\\
\theta_i^c=&\mathbb{E}_{q(\omega_i^c,n_i^c)}\left[\omega_i^c\right] = \frac{{y'}_i^c+\gamma_i^c}{2b_i^c}\tanh{\frac{b_i^c}{2}}\nonumber\\
\bmu^c =&  \frac{1}{2}({\Sigma^c})^{-1}{\kappa^c}^\top\left({\by'}^c-\bgamma^c\right) \label{eq:global_updates_1}\\
\Sigma^c =& \left({\kappa^c}^\top \text{diag}\left(\boldsymbol{\theta}^c\right)\kappa^c+({K_{mm}^{c}})^{-1}\right)^{-1}\label{eq:global_updates_2},
\end{align}
where $\psi(.)$ is the digamma function. 
When $\kappa\mu \ll 0$, equation~\eqref{eq:local_updates_gamma} easily overflows. One can solve this problem by approximating $\exp(-0.5\kappa\mu)/\cosh(0.5\bar{f})$ with $\sigma(\kappa\mu)$ by neglecting the variance terms $\widetilde{K}+\kappa\Sigma{\kappa}^\top$ in $\bar{f}$.

Equation~\eqref{eq:local_updates_gamma} and \eqref{eq:local_updates_alpha} shows a direct interdependence between $\alpha_i$ and $\gamma_i^c$. 
We use inner loop of alternating between updating both variables until convergence to solve the problem. We find that 5 iterations in the inner loop are enough.
\begin{figure*}
	\centering
	\includegraphics[width=\textwidth]{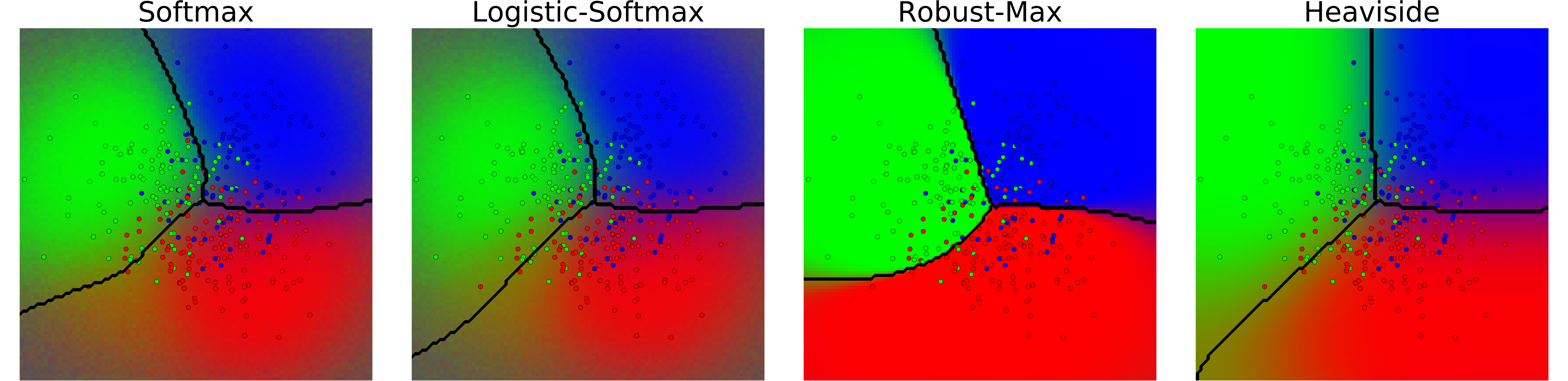}
	\caption{RGB representation of the predictive likelihood for a toy dataset as described in section~\ref{sec:exp_comp} with variance $\sigma^2=0.5$. Each class is attributed a color channel (Red, Green, Blue) and predictive likelihoods are mapped into RGB values.}
	\label{fig:contour_05}
\end{figure*}

Finally, if class subsampling (the extreme classification version of our algorithm Alg.~\ref{algo2}) is used, $\alpha_i$ is approximated by
\begin{equation}
\alpha_i = 1 + \frac{C}{|\mathcal{K}|}\sum_{c\in \mathcal{K}} \gamma_i^c,
\label{eq:noisy_updates_alpha}
\end{equation}
where $C$ is the number of classes and $|\mathcal{K}|$ is the number of sub-sampled classes.

\subsection{Subsampling the classes (extreme classification version)}
\label{appendix:subclasses}
The extreme classification version of our algorithm is presented in Alg.~\ref{algo2}.
In each iteration we only consider a minibatch of the classes $\mathcal B \subset \{1,\dots,C\}$ and the variational parameters $b_i^c$, $\alpha_i^c$, $\mu^c$, $\Sigma^c$ (lines \ref{algo:line_local}, \ref{algo:line_alpha_lambda}, \ref{algo:line_global1}, \ref{algo:line_global2} in Alg.~\ref{algo}) are only updated for $i \in \mathcal B$. The updates that are global w.r.t. the classes, i.e. $\lambda_i$ and the hyperparameters $h$ (lines \ref{algo:line_alpha_lambda}, \ref{algo:line_hyper}) are now replaced by stochastic gradient updates.

\begin{algorithm}[H]
	\caption{\methodname~with class subsampling}
	\small
	\begin{algorithmic}[1]
		\State \textbf{Input:} data $\bX$,$\by$, minibatch size $|\mathcal S|$and $|\mathcal B|$
		\State \textbf{Output:} variational posterior GPs $p(u^c | \mu^c, \Sigma^c)$
		\State Set the learning rate schedules $\rho_t, \rho^h_t$ appropriately
		\State Initialize all variational parameters and hyperparameters
		\State Select $M$ inducing points locations (e.g. kMeans)
		\For {iteration $t=1,2,\dots$}
		\State {\tt \# Sample minibatch:}
		\State Sample a minibatch of the data ${\mathcal S} \subset \{1,\dots,N\}$
		\State Sample a set of labels ${\mathcal K} \subset \{1,\dots,C\}$
		
		\State {\tt \# Local variational updates}
		\For {$i \in \mathcal S$}
		\State Update $(\alpha_i, \gamma_i^c)_{c\in\mathcal{K}}$ \label{algo2:line_alpha_lambda} (Eq.~\ref{eq:local_updates_gamma},\ref{eq:noisy_updates_alpha})
		\For { $c \in \mathcal K$}
		\State Update $b_i^c$ (Eq.~\ref{eq:local_updates_omega}) \label{algo2:line_local}
		\EndFor
		\EndFor
		
		\State {\tt \# Global variational GP updates}
		\For {$c \in \mathcal{K}$}
		\State $\mu^c \gets (1-\rho_t) \mu^c + \rho_t \hat \mu^c $ (Eq.~\ref{eq:global_updates_1}) \label{algo2:line_global1}
		\State $\Sigma^c \gets (1-\rho_t) \Sigma^c + \rho_t \hat \Sigma^c $ (Eq.~\ref{eq:global_updates_2}) \label{algo2:line_global2}
		\EndFor
		\State {\tt \# Hyperparameter updates}
		\State Gradient step $h \gets h + \rho^h_t \nabla_h \mathcal L$ \label{algo2:line_hyper}
		\EndFor
		
	\end{algorithmic}
	\label{algo2}
\end{algorithm}

\subsection{Visualization of the different likelihoods}
\label{appendix:decision_boundary}
To get a better intuition of the behavior of each likelihood, we visualize the prediction function of each method as a contour plot using the toy dataset from section~\ref{sec:exp_comp}. To visualize the predictive likelihood, we map the predictive values of each class to a RGB color channel (where each class corresponds to one color and mixing of colors indicates a contribution of multiple classes). A highly saturated color corresponds to a high confidence in the class prediction, while mixed colors indicate zones of transition between classes and lower confidence. The results are shown in Figure~\ref{fig:contour_05} for a toy dataset consisting of 500 points generated from a mixture of Gaussians with variance $\sigma^2=0.5$. As expected, the robust-max likelihood leads to extremely sharp decision boundaries and high confidences for all regions (even for the overlapping regions). The other likelihoods lead to better calibration resulting in soft boundaries and less confident predictions in the overlapping regions.

\subsection{Convexity of the negative ELBO}
\label{appendix:convex}
In the following we prove that the negative ELBO ($-\mathcal{L}$) of our augmented model is convex in the global variational parameters $\mu^c$ and $\Sigma^c$. To prove this statement, we write the negative ELBO in terms of $\mu^c$ and $\Sigma^c$,
\begin{align}
-\mathcal{L}(\mu^c,\Sigma^c) \uptoconst& \frac{1}{2}\left[ \sum_{i=1}^N (y'^c_i-\gamma_i^c)\mu_i^c-\theta^c_i \left((\mu_i^c)^2+\Sigma_{ii}^c\right)\right]\nonumber\\
& \frac{1}{2}\left[{\mu^c}^\top K^{-1} \mu^c + \text{tr}(K^{-1}\Sigma^c) - \log|\Sigma^c|\right].\nonumber
\end{align}
Differentiating twice in $\mu^c$ gives $\text{diag}(\theta^c) + K^{-1}$ which is positive definite since $\theta^c_i > 0$ for all $i$ and by definition of $K$. Therefore, the negative ELBO is convex in $\mu^c$ for all $c$.

Differentiating twice in $\Sigma^c$ gives $\left(\Sigma^c\right)^{-1}\otimes\left(\Sigma^c\right)^{-1}$, where $\otimes$ is the Kroenecker product. This is again positive definite since $\left(\Sigma^c\right)^{-1}$ is positive definite and the Kroenecker product preserves positive definiteness. Therefore, the negative ELBO is also convex in $\Sigma^c$ for all $c$.



\end{document}